\documentclass[runningheads]{llncs}
\usepackage{arxiv}

\usepackage{graphicx}
\usepackage{hyphenat}
\usepackage{multirow}
\usepackage[frozencache,cachedir=.]{minted}

\usepackage{caption}
\usepackage{subcaption}
\usepackage{xurl}

\usepackage{listings}

\begin{document}
\title{WDV: A Broad Data Verbalisation Dataset Built from Wikidata}
\titlerunning{WDV: A Wikidata Verbalisation Dataset}

\author{Gabriel Amaral\inst{1}\orcidID{0000-0002-4482-5376} \and
{Odinaldo Rodrigues \inst{1}\orcidID{0000-0001-7823-1034}} \and
Elena Simperl \inst{1}\orcidID{0000-0003-1722-947X}}

\authorrunning{G. Amaral et al.}
%
\institute{King's College London, London WC2R 2LS, UK\\
\email{\{gabriel.amaral,odinaldo.rodrigues,elena.simperl\}@kcl.ac.uk}}
\maketitle              
\begin{abstract} 

Data verbalisation is a task of great importance in the current field of natural language processing, as there is great benefit in the transformation of our abundant structured and semi-structured data into human-readable formats. Verbalising Knowledge Graph (KG) data focuses on converting interconnected triple-based claims, formed of subject, predicate, and object, into text. Although KG verbalisation datasets exist for some KGs, there are still gaps in their fitness for use in many scenarios. This is especially true for Wikidata, where available datasets either loosely couple claim sets with textual information or heavily focus on predicates around biographies, cities, and countries. 
To address these gaps, we propose WDV, a large KG claim verbalisation dataset built from Wikidata, with a tight coupling between triples and text, covering a wide variety of entities and predicates. We also evaluate the quality of our verbalisations through a reusable workflow for measuring human-centred fluency and adequacy scores. Our data\footnote{\url{https://doi.org/10.6084/m9.figshare.17159045.v1}} and code\footnote{\url{https://github.com/gabrielmaia7/WDV}} are openly available in the hopes of furthering research towards KG verbalisation.

\end{abstract}  
\section{Introduction} 
\label{sec:introduction}

Data verbalisation, a facet of Natural Language Generation (NLG), is a task that has great importance in the current field of natural language processing~\cite{ribeiro-etal-2021-investigating,harkous-etal-2020-text,guo-etal-2020-2,trisedya-etal-2018-gtr,zhao-etal-2020-bridging,ijcai2020-419}, as there is great benefit in the transformation of our abundant structured and semi-structured data into human-readable formats. It is important in its own right, as well as as a step towards larger tasks such as open-domain question-answering~\cite{ma2021open} and automated fact checking~\cite{yang-etal-2020-program,yang2021exploring}. One large source of semi-structured data that would benefit greatly from verbalisation is collaborative Knowledge Graphs (KG) like DBpedia\footnote{\url{https://www.dbpedia.org/}} and Wikidata.\footnote{\url{https://www.wikidata.org}}

The verbalisation of KGs data consists of converting sets of claims into natural language text. Each claim consists of a triple, formed of subject, predicate, and object, and each claim set shares subjects and objects; the verbalisation then has to deal with expressing and linking these pieces of information. Although KG verbalisation datasets, mapping claim sets to text, exist for some popular KGs~\cite{gardent-etal-2017-webnlg,bosc-etal-2016-dart,elsahar-etal-2018-rex}, they are not without their limitations. 

Wikidata, the web's largest collaborative KG, has very few such datasets~\cite{ijcai2020-711,elsahar-etal-2018-rex}, and existing ones rely on distant supervision to prioritise the sheer number of couplings in exchange for coupling tightness. In addition, they disproportionately represent specific entity types from Wikidata, such as people and locations, when Wikidata covers a much wider variety of information.

Finally, data verbalisation performance is mainly measured with algorithmic approaches, such as BLEU~\cite{bleu} and METEOR~\cite{banerjee-lavie-2005-meteor}, which have been the target of many criticisms when applied to NLG~\cite{reiter2018structured,novikova-etal-2017-need,sulem-etal-2018-bleu}.
To address these gaps, we propose WDV, a large KG verbalisation dataset with 7.6k entries extracted from Wikidata. Our contributions are threefold:
\begin{enumerate}
    \item WDV is built from a much wider variety of entity types and predicates than similar datasets, and is intended as a benchmarking dataset for data verbalisation models applied on Wikidata;
    \item WDV supports a tight coupling between single claims and text directly associating a triple-based claim and a natural language sentence;
    \item 1.4k entries of WDV have been annotated by a collective of humans, allowing for the evaluation and future improvement of our verbalisations, as well as to establish a non-algorithmic baseline for other verbalisation models.
\end{enumerate}

Additionally, we create a reproducible crowdsourcing workflow for capturing human evaluations of fluency and adequacy in graph-to-text NLG. All used code and gathered data is available in this paper's GitHub repository. 

%
\section{Background and Related Work} 
\label{sec:related_work}

Verbalising KGs consists of generating grammatically correct natural language based on structured and semi-structured data from a KG, maintaining original meaning. This data is encoded in triples (claims), consisting of a subject, a predicate, and an object; all three components model aspects of knowledge, such as entities, classes, attributes, and relationships. Examples of popular KGs are DBpedia, Wikidata, Yago,\footnote{\url{https://github.com/yago-naga/yago3}} and Freebase.\footnote{\url{https://developers.google.com/freebase}} Their verbalisation is an important task on its own, but is also a key step in downstream tasks~\cite{ma2021open,yang-etal-2020-program,yang2021exploring,vlachos-riedel-2015-identification}.

Datasets that align KG claims to text are vital for creating and evaluating KG verbalisation approaches. While several have been created, they are not without their limitations.
The \textit{NYT-FB}~\cite{mintz-etal-2009-distant,yao-etal-2011-structured} dataset aligns text from the New York Times with triples from Freebase through named entity linking and keyword matching against Freebase labels. This leads to a disproportional coverage of news-worthy entities and topics, such as geography and politics, and from a specific period in time, limiting its usefulness on broader scenarios.
The same narrow scope is seen in the \textit{TACRED} dataset~\cite{zhang-etal-2017-position}, which covers only 41 relationships about people and organisations, such as age, spouse, shareholders, etc, as its data does not stem from any specific KG, but rather annotated newswire and web text from the TAC KBP corpora~\cite{AB2/VC89SM_2018}. Also, its texts often contain much more information than their aligned triples, making it a resource not fully suited for NLG.
The \textit{FB15K-237} dataset~\cite{Toutanova2015ObservedVL} aligns Freebase triples to synsets instead of text, making it unusable for NLG without text grounding. Additionally, both NYT-FB and FB15K-237 rely on Freebase, which was discontinued and its data moved to Wikidata~\cite{10.1145/2872427.2874809}, compromising these datasets' usability and upkeep.

More recent datasets attempt to remedy some of these limitations.
Pavlos et al.~\cite{vougiouklis2017neural,ijcai2020-711} propose two large corpora that align Wikidata and DBpedia claims to Wikipedia text. However, they focus on verbalisations of multiple claims at a time, which limits its usefulness for important tasks e.g. automated fact-checking in favour of others e.g. summarisation. Even more critically, they are based on distant supervision techniques, providing a loose alignment between sets of triples and text; triple sets consist of numerous claims that are very likely — but not guaranteed — to be expressed in the text, and the text contains information that is not assured to exist in the claims. The same is true for \textit{T-REx}~\cite{elsahar-etal-2018-rex}, which aligns Wikidata claims to Wikipedia abstracts, making it unreliable for NLG from KG claims while perfectly preserving their sense. Our dataset bridges this gap by focusing on a tight alignment between Wikidata claims and text.

The coverage issue seen in NYT-FB and TACRED is also present, although less so, in T-REx. It covers many unique predicates, yet they are disproportionately represented: the top 7.7\% of its unique predicates represent $90\%$ of its unique triples, and these mostly express information on people and places — the \texttt{country} predicate alone represents over $11\%$ of triples. The \textit{WebNLG}~\cite{gardent-etal-2017-webnlg} dataset remedies this by defining a list of very broad DBpedia classes and then collecting separate and balanced sets of claims from entities in each class. However, WebNLG also focuses on sets of multiple claims at a time.

We follow WebNLG's approach to resolving predicate and theme bias. However, we build WDV out of Wikidata instead, expanding the entity classes defined by WebNLG, as Wikidata lacks verbalisation datasets that cover its wide range of predicates and themes. 
To provide a better view of how WDV compares to other datasets mentioned in this Section, refer to Table~\ref{tab:datasets}.

\begin{table}[ht]
\centering
{\small
\setlength{\tabcolsep}{3pt}
\renewcommand{\arraystretch}{0.9}
\vspace*{-8pt}
\begin{tabular}{l|l|l|l|l|l|l|l|}
\cline{2-8}
                                   & \begin{tabular}[c]{@{}l@{}}Source\\ Graph\end{tabular} & \begin{tabular}[c]{@{}l@{}}Aligned\\ Documents\end{tabular} & \begin{tabular}[c]{@{}l@{}}Unique\\ Predicates\end{tabular} & \begin{tabular}[c]{@{}l@{}}Unique\\ Triples\end{tabular} & \begin{tabular}[c]{@{}l@{}}Entity\\ Classes\end{tabular} & \begin{tabular}[c]{@{}l@{}}Text\\ Alignment\end{tabular} & Avail. \\ \hline
\multicolumn{1}{|l|}{NYT-FB}       & Freebase                                               & 1.8M                                                        & 258                                                         & 39K                                                      & n.a.                                                     & Distant                                                  & Partial      \\ \hline
\multicolumn{1}{|l|}{TACRED}       & n.a.                                                   & 106K                                                        & 41                                                          & 21K                                                      & n.a.                                                     & Distant                                                  & Closed       \\ \hline
\multicolumn{1}{|l|}{FB15K-237}    & Freebase                                               & 2.7M                                                        & 237                                                         & 2.7M                                                     & n.a.                                                     & Tight                                       & Public       \\ \hline
\multicolumn{1}{|l|}{T-REx}        & Wikidata                                               & 6.2M                                                        & 642                                                         & 11M                                                      & n.a.                                                     & Distant                                                  & Public       \\ \hline
\multicolumn{1}{|l|}{WebNLG}       & DBpedia                                                & 39K                                                         & 412                                                         & 3.2K                                                     & 16                                                       & Tight                                                   & Public       \\ \hline
\multicolumn{1}{|l|}{\textbf{WDV}} & Wikidata                                               & 7.6K                                                        & 439                                                         & 7.6K                                                     & 20                                                       & Tight                                                   & Public       \\ \hline
\end{tabular}
}
\vspace*{10pt}
\caption{Comparison between WDV and other KG verbalisation datasets. `Entity Classes' shows in how many distinct themes the claims might be organised by, if at all. `Text Alignment' refers to whether all text corresponds to aligned triples (Tight) or not (Distant). \textit{Avail.} stands for Availability.}
\label{tab:datasets}
\vspace*{-24pt}
\end{table}

\section{WDV: An Annotated Wikidata Verbalisation Dataset} 
\label{sec:dataset}

\begin{figure}[ht]
  \centering
  \vspace*{-15pt}
  \includegraphics[width=1\linewidth]{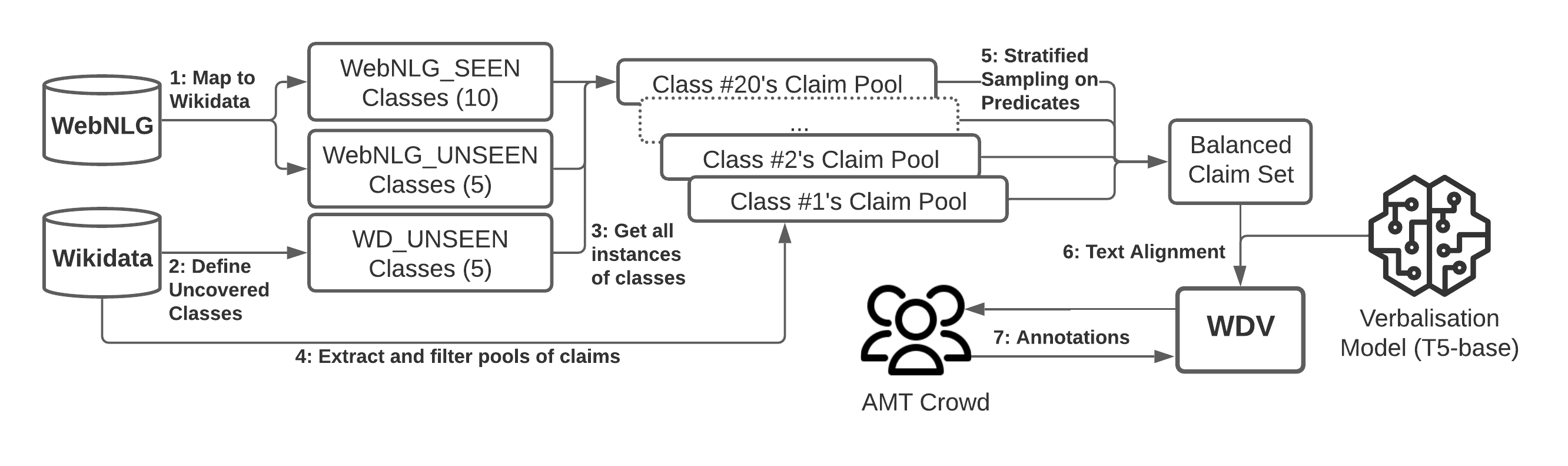}
  \vspace*{-15pt}
  \caption{Overview of WDV's construction workflow, starting with WebNLG's structure and Wikidata's contents, finishing with WDV and crowd annotations.}
  \label{fig:workflow}
\vspace*{-24pt}
\end{figure}

This section describes the construction of the WDV dataset, including crowdsourced annotations carried, as well as details of its structure. Figure~\ref{fig:workflow} illustrates the entire process with numbered steps, which we cover in this Section. In a nutshell, it consists of first defining 20 large pools of filtered Wikidata claims, each corresponding to a Wikidata class (steps 1-4). Then, we obtain a sample of claims from each pool such that predicates are represented as equally as possible (step 5). Lastly, we obtain aligned verbalisations and human annotations (steps 6 and 7). Throughout this entire construction process, data was extracted from a Wikipedia JSON dump from August 2021. The JSON format was used since the later stages of the pipeline i.e. crowdsourcing and verbalisation either require or greatly benefit from that input format. We also release WDV in this format as it targets ML practitioners and developers, who are very familiar with it.

To improve comprehensibility, transparency, and repeatability, we follow two recently proposed sets of guidelines. The first, by Gebru et al.~\cite{gebru2020datasheets}, pertains to the effective documentation of machine learning datasets, supporting the transparency and reproducibility of their creation process. The second, by Ramirez et al.~\cite{ramirez2021state}, pertains to the detailing of crowdsourcing experiments to guarantee clarity and repeatability. It ensures the impact of task design, data processing, and other factors on our conclusions, as well as their validity, can be assessed.

\subsection{Balanced Claim Set Collection}
\label{sec:dataset:collection}

WDV adapts and expands on WebNLG's partition and class structure to fit Wikidata. Firstly, this ensures a balanced representation of Wikidata entities and predicates of various natures. Secondly, our data verbalisation model, used later in the workflow, is fine-tuned with WebNLG; keeping the same class composition thus reduces the chances of low-quality verbalisations. WebNLG has two partitions: \textit{SEEN}, with 15 classes, and \textit{UNSEEN}, with five, as seen in Table~\ref{tab:dataset_breakdown}.

We start by mapping WebNLG's 15 DBpedia classes to their Wikidata equivalents (\textbf{step 1}). 
Some of Wikidata's most populous classes are not at all covered by these 15. Thus, from these uncovered classes, we select the five largest to compose an additional partition \textit{WD\_UNSEEN} (\textbf{step 2}); we do not consider ontological or scientifically complex classes (e.g. proteins). Next, we extract from Wikidata all entities that can be considered as instances or examples of these 20 classes or their subclasses (\textbf{step 3}), composing 20 large groups of entities. 

From each class' extracted group of entities, we retrieve all claims that we deem suitable for verbalisation, based on the following exclusion criteria (\textbf{step 4}): we exclude deprecated claims, as they might contain incorrect or invalid values; claims with objects of datatypes that are hard to represent in natural language are excluded e.g. external database identifiers, URLs, images, mathematical formulas, etc; we exclude claims that serve taxonomic or ontological purposes e.g. \texttt{subclass of (P31)}, \texttt{Topic's main category (P910)}, \texttt{See also (P1659)} etc; and finally, claims whose objects are the special values \textit{no value} or \textit{some value}. The claims remaining after these exclusions compose 20 distinct pools of claims, or \textit{themes}, from which we will next sample a set of claims.

These themes have very unbalanced distributions of claims over predicates e.g. over $50\%$ of the claims in the \textit{Airport} and \textit{Mountain} themes have the \texttt{patronage (P3872)} and \texttt{country (P17)} predicates, respectively. A simple random sample would build a dataset that ignores the vast majority of Wikidata predicates. Hence, we opt for a stratified sampling of claims (\textbf{step 5}). 

For each theme \texttt{t}, we determine the representative sample size \texttt{$N_t$} needed, considering its total number of claims, a $95\%$ confidence interval, and a $5\%$ margin of error. We start the sampling process by grouping each theme's claims by predicate, discarding very rare predicates ($0.3\%$ to $1.7\%$ of total claims in a theme), and defining each theme's remaining \texttt{$M_t$} predicate groups as a stratum. For each theme \texttt{t}, we attempt to sample an equal amount of claims (\texttt{$N_{t}/M_{t}$}) from each stratum. If a stratum in theme \texttt{t} has less than \texttt{$N_{t}/M_{t}$} claims, we select all its claims and compensate by oversampling other strata in \texttt{t}, so that total sample size is still \texttt{$N_t$}. We keep track of all sampling weights in order to adjust any estimated statistic to account for the stratification. The resulting balanced claim set consists of statistically representative sets of claims from all 20 themes (7.6k claims in total), where predicates are as equally present as possible.

\subsection{Text Alignment}
\label{sec:dataset:text}
WDV tightly aligns each claim to a natural language text i.e. each claim corresponds exactly to one sentence (and vice-versa), such that both hold the same meaning and the sentence is grammatically well-written. This is so that NLG is directly supported (as explored in Section~\ref{sec:related_work}, and also because WDV is the first step towards future research into automating AFC for Wikidata. 

To achieve this alignment (\textbf{step 6}), we first collect subject, predicate, and object labels (preferably in English) for each claim in the balanced claim set. We also collect aliases and descriptions, which play a part later in crowdsourcing. The collection is done by querying Wikidata's SPARQL engine.\footnote{\url{https://query.wikidata.org/}} In cases such as timestamps and measurements with units, label templates are used.

For each claim, its three labels are given to a verbalisation model, which outputs an English sentence that attempts to communicate the same information. The model itself, including its training and validation, is detailed in Section~\ref{sec:verbalisation}. This results in 7.6k claim-verbalisation pairings. 

These claim-verbalisation pairings, alongside ontological attributes and the aggregated crowdsourced annotations (see Section~\ref{sec:crowdsourcing}), constitute WDV. Its detailed structure, an exemplary record, and some descriptive statistics are given in Section~\ref{sec:dataset:composition}. Section~\ref{sec:data_analysis} explores insights obtained from crowd annotations.

\subsection{Crowdsourced Annotations} 
\label{sec:crowdsourcing}

To measure how much of the claims' meanings are kept (i.e. adequacy) by the verbalisations and how much they resemble text written by humans (i.e. fluency), as well as to support the dataset's refining and correction, we crowdsource human annotations (\textbf{step 7}). These annotations are collected for a portion of WDV ($20\%$ of total claims) due to budget constraints, randomly selected among those claims having all labels in English, while keeping a proportional representation of each theme. Claim components not labelled in English are a minority that would represent a hurdle for crowd workers~\cite{DBLP:journals/csl/LeeGMK21} and bias results.

\subsubsection{Experimental Design}

Before crowdsourcing, the WDV data goes through two pre-processing steps: \textit{golden data generation} and \textit{task composition}. Golden data is a small data subset that is manually annotated and used as a reference to discern between good and bad workers. We calculate how much golden data is necessary by minimizing, based on available data from similar studies~\cite{10.1145/3484828}, the probability of a regular worker finding a repeated set of golden data in two different tasks, which plateaus near $100\%$ with $90$ golden data annotations.

We take $45$ random records from the sampled WDV data and set them aside as golden data for both fluency and adequacy tasks. We manually generate another $90$ uniquely identified pairs to represent poor model performance: $45$ for the fluency task by writing different levels of gibberish, and $45$ for adequacy by randomly shuffling their verbalisations. We annotate golden data by defining, for each pair, what would represent reasonable scores for fluency and adequacy.

Task composition consists of: first, grouping the sampled WDV data plus the golden data such that each group (a \textit{task set}) has two random golden data pairs and four random non-annotated pairs; then, attributing to each task a unique identifier; and lastly, sending the task set to the crowd embedded in an HTML script to be solved by at least five different workers.

Pilots were run in August 2021, and main tasks were run between September and October of the same year. Pilots helped us measure median time spent by workers to define fair payment, and collect general feedback to adjust task design. We calculated pay based on the double of US's minimum hourly wage of USD$7.25$, in order to properly account for good workers that need more time than the median. We paid USD$0.50$ per fluency task and USD$1.00$ per adequacy task. Workers rated our tasks as having fair pay on TurkerView.\footnote{\url{https://turkerview.com/}} Before starting the task, workers are made aware of the pay and conditions and are told that continuing with the task means consenting to both.

\subsubsection{Crowd}

Crowd workers were selected from AMT's pool of workers, the demographics of which have been explored in several papers~\cite{djellel2018demographics,huff2015evaluating,burnham2018personal}. We limited the tasks only to workers that had a good grasp of English by including an English grammar screening quiz before each task. Secondly, we only allowed workers that had done over 1000 tasks with over 80\% acceptance rate to work on our tasks. We analysed contributions from the pilot, identifying workers that exhibited malicious behaviour and banning them from the main tasks. 

\subsubsection{Tasks}

Task sets are sent to be annotated embedded in HTML pages. There is one for \textit{fluency} and one for \textit{adequacy} annotation tasks. Before starting either task type, workers are shown a description of that task, rules, and instructions they should follow. They also see many examples of acceptable answers with explanations. Workers can access this information at all times during the task.

In the fluency task, workers are shown only the verbalisation and are asked to rate its fluency with a score from 0 to 5, 0 being the worst and 5 being the best. In the adequacy task, workers are shown both the verbalisation and the claim, as well as labels, aliases, and descriptions, and are asked whether they convey the same information. They can reply \textit{Yes} (giving it a score of 0), \textit{No} (score of 1), and \textit{Not Sure} (score of 2). Answering \textit{No} 
and \textit{Not Sure} prompts a question as to the reason; workers can blame the verbalisation, each component in the triple, a combination, or select \textit{Other} and give a new justification. These tasks were released on AMT after receiving ethical approval.

\subsubsection{Quality Control}

Multiple quality control techniques were applied. The small randomized grammar quiz at the start of the task serves as an attention check, discouraging spammers. Our gold data is used to measure worker quality during the task, alongside other checks such as time spent per pair and whether all questions were answered. Failing these checks alerts the user and asks them to reevaluate their annotations. Failing three times closes the task without submission. Workers are told these details before engaging with the task.


\subsubsection{Task Code and Raw Data}

All the code and data for our crowdsourcing is in this paper's GitHub repository, including detailed descriptions of each task's execution and the exact HTML code sent to each anonymous worker alongside instructions, agreement terms, and examples. It also includes all retrieved data before it was processed and aggregated back into WDV. 

\subsection{WDV Composition}
\label{sec:dataset:composition}

WDV consists of a large partially annotated dataset of over 7.6k entries that align a broad collection of Wikidata claims with their respective verbalisations. An example of an annotated record can be seen in Listing~\ref{lst:json-example}. The attributes seen there consist of: attributes describing the claim, such as its Wikidata ID (\textit{claim\_id}) and its \textit{rank} (normal, deprecated or preferred); attributes from the claim's components (subject, predicate, and object), including their Wikidata IDs (e.g. \textit{subject\_id}), labels (e.g. \textit{subject\_label}), descriptions (e.g. \textit{subject\_desc}), and aliases (e.g. \textit{subject\_alias}); a JSON representation of the \textit{object} alongside its type (\textit{object\_datatype)} as defined by Wikidata; attributes from the claim's theme such as its root class' Wikidata ID (theme\_root\_class\_id) and label (theme\_label); the aligned \textit{verbalisation}, before and after replacement of tokens unknown to the model (\textit{verbalisation\_unk\_replaced}); the \textit{sampling weight} from the stratified sampling process; and the crowdsourced \textit{annotations} and their aggregations, for those entries ($\sim$1.4k) that are annotated.

Our schema is different from the Wikipedia dumps' JSON schema. Firstly, the latter is entity-centered: each entry is an entity and claims are components hierarchically encoded as elements. As WDV is centered on claim-verbalisation alignments, we flatten this structure. Secondly, information on the claims' components is spread over their respective JSON objects. Our schema organises all relevant data about the claim-verbalisation pair in a single JSON object.

WDV is a 3 star dataset according to the 5 star deployment scheme for Linked Data.\footnote{\url{https://www.w3.org/2011/gld/wiki/5_Star_Linked_Data}} It is available on the web in a structured, machine-readable, and non-proprietary format. Making it 4 star by converting it into RDF is our immediate next step. Wikidata already has a well-documented RDF representation schema,\footnote{\url{https://www.mediawiki.org/wiki/Wikibase/Indexing/RDF_Dump_Format}} reified based on n-ary relationships~\cite{10.1007/978-3-319-11964-9_4}. We will make use of this schema to express the data about the claim and its components (e.g. ids, rank, labels, descriptions, values, etc.), as they are already explicitly supported by it, and it is an effective way to represent Wikidata in RDF~\cite{Hernndez2015ReifyingRW}. We will then complement it with custom vocabulary in order to express the verbalisations and their crowdsourced annotations. We can do this by linking the statements, expressed in Wikidata's RDF schema as nodes, to a verbalisation node through a \texttt{wdv:verbalisation} predicate, which then is linked to its crowdsourced annotations through fitting predicates, e.g. \texttt{wdv:fluencyScore} and \texttt{wdv:adequacyScore}. We can also reuse existing vocabularies, such as LIME~\cite{10.1007/978-3-319-18818-8_20}).

\vspace{15pt}
\begin{listing}[ht]
{\small
\vspace*{-12pt}
\begin{minted}[breaklines=true,fontsize=\small,baselinestretch=1]{js}
{   "claim_id": "Q55425899$D1CB6CEC-33E4-41DF-9244-3277C2BE1FA5"
    "rank" : "normal",
    "subject_id" : "Q55425899",
    "property_id" : "P6216",
    "subject_label" : "Spring in Jølster",
    "property_label" : "copyright status",
    "object_label" : "public domain",
    "subject_desc" : "painting by Nikolai Astrup",
    "property_desc" : "copyright status for intellectual creations like works of art, publications, software, etc.",
    "object_desc" : "works that are no longer in copyright term or were never protected by copyright law",
    "subject_alias" : "no-alias",
    "property_alias" : ["copyright restriction"],
    "object_alias" : ["PD", "out of copyright", "DP"],
    "object_datatype" : "wikibase-item",
    "object" : { "value": {"entity-type": "item", "numeric-id": 19652, "id": 'Q19652'},
                 "type": "wikibase-entityid" },
    "theme_root_class_id" : "Q3305213",
    "theme_label" : "Painting",
    "verbalisation" : "Spring in J <unk> lster is in the public domain.",
    "verbalisation_unk_replaced" : "Spring in Jølster is in the public domain.",
    "sampling_weight" : 3538.615384615385,
    "annotations": { "fluency_scores" : [5, 4, 4, 2, 1],
                     "fluency_mean" : 3.2,
                     "fluency_median" : 4.0,
                     "adequacy_scores" : [0, 0, 1, 0, 0],
                     "adequacy_majority_voted" : 0,
                     "adequacy_percentage" : 0.8 }
}
\end{minted}
\vspace*{-12pt}
\caption{Example of an annotated record from WDV in JSON format\label{lst:json-example}} 
}
\end{listing}
\vspace*{-12pt}

Table~\ref{tab:dataset_breakdown} shows a breakdown of WDV. In the first column, we can identify the \textit{SEEN} and \textit{UNSEEN} partitions from WebNLG, as well as our added \textit{WD\_UNSEEN} partition built from other Wikidata classes. The second column divides them into component themes (or pools of claims). For each theme, it then shows the number of unique properties (predicates), unique claims (calculated as \texttt{$N_t$}, as described in Section~\ref{sec:dataset:collection}), and how many were annotated.

\begin{table}[ht]
\centering
{
\vspace*{0pt}
\begin{tabular}{|c|l|r|r|r|}
\hline
Partition                       & \multicolumn{1}{c|}{Theme} & \multicolumn{1}{c|}{Properties} & \multicolumn{1}{c|}{Claims} & \multicolumn{1}{c|}{Annotated Claims} \\ \hline
\multirow{10}{*}{WebNLG\_SEEN}  & Airport                    & 27                              & 382                         & 76                                    \\ \cline{2-5} 
                                & Astronaut                  & 57                              & 351                         & 71                                    \\ \cline{2-5} 
                                & Building                   & 67                              & 385                         & 63                                    \\ \cline{2-5} 
                                & City                       & 72                              & 383                         & 73                                    \\ \cline{2-5} 
                                & ComicsCharacter            & 79                              & 376                         & 76                                    \\ \cline{2-5} 
                                & Food                       & 64                              & 368                         & 67                                    \\ \cline{2-5} 
                                & Monument                   & 62                              & 380                         & 51                                    \\ \cline{2-5} 
                                & SportsTeam                 & 49                              & 383                         & 75                                    \\ \cline{2-5} 
                                & University                 & 62                              & 378                         & 75                                    \\ \cline{2-5} 
                                & WrittenWork                & 21                              & 385                         & 66                                    \\ \hline
\multirow{5}{*}{WebNLG\_UNSEEN} & Artist                     & 65                              & 384                         & 78                                    \\ \cline{2-5} 
                                & Athlete                    & 53                              & 385                         & 80                                    \\ \cline{2-5} 
                                & CelestialBody              & 25                              & 385                         & 83                                    \\ \cline{2-5} 
                                & MeanOfTransportation       & 58                              & 376                         & 71                                    \\ \cline{2-5} 
                                & Politician                 & 56                              & 385                         & 75                                    \\ \hline
\multirow{5}{*}{WD\_UNSEEN}     & ChemicalCompound           & 33                              & 383                         & 81                                    \\ \cline{2-5} 
                                & Mountain                   & 23                              & 380                         & 69                                    \\ \cline{2-5} 
                                & Painting                   & 29                              & 385                         & 50                                    \\ \cline{2-5} 
                                & Street                     & 21                              & 384                         & 66                                    \\ \cline{2-5} 
                                & Taxon                      & 27                              & 385                         & 80                                    \\ \hline
ALL                             & ALL                        & 439                             & 7607                        & 1426                                  \\ \hline
\end{tabular}
}
\vspace*{10pt}
\caption{Total number of unique properties, unique claims, and annotated claims, per partition and themes in WDV.}
\label{tab:dataset_breakdown}
\vspace*{-24pt}
\end{table}

\subsection{Crowd Data and Risk Analysis} 
\label{sec:data_analysis}

Crowdsourced annotations were aggregated and added to WDV as attributes, as depicted in Section~\ref{sec:dataset:composition}. In this section, we analyse these aggregated annotations and draw conclusions on the quality and reliability of WDV.

\subsubsection{Aggregation and Reliability}
\label{sec:data_analysis:agg}

Fluency scores were aggregated by calculating both median and mean, in case more or less weight, respectively, needs to be given to workers who disagree greatly with their peers. Adequacy was aggregated by majority voting, and also by calculating the percentage of workers that voted \textit{Yes}, which we call \textit{adequacy percentage}.

Fluency has been fair to very high in most verbalisations. A fluency score of 3 indicates ``Comprehensible text with minor grammatical errors'', and over $96\%$ of verbalisations find themselves with median fluency equal to or above 3. This shows our verbalisation model produces fluent text from Wikidata triples. The model also maintains very well the meaning of Wikidata claims after verbalising. Almost $93\%$ of verbalisations are majority-voted as adequate.

The reliability of aggregated crowdsourced data can be indicated by statistical inter-annotator agreement metrics~\cite{nowak2010reliable} such as Krippendorff's Alpha~\cite{krippendorff}. The alpha measured for the fluency scores is $0.4272$, and for the adequacy scores it is $0.4583$; both indicate moderate agreement, according to the interpretations recommended by Landis \& Koch~\cite{landis1977measurement}.

\subsubsection{Variations in Scores and Agreement}
Next, we see how fluency, adequacy, and agreement might vary across the partitions and themes shown in Table~\ref{tab:dataset_breakdown}.

\begin{figure}[ht]
\centering
\begin{subfigure}{.45\textwidth}
  \centering
  \includegraphics[width=1\linewidth]{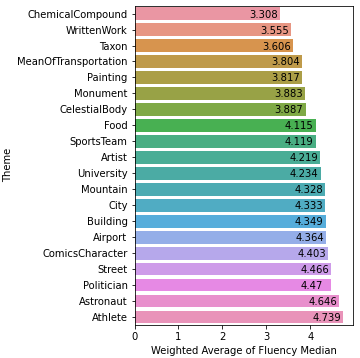}
  \caption{\label{fig:weighted_aggregated_flu}}
\end{subfigure}%
\begin{subfigure}{.45\textwidth}
  \centering
  \includegraphics[width=1\linewidth]{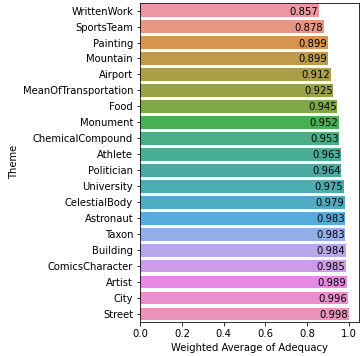}
  \caption{\label{fig:weighted_aggregated_ade}}
\end{subfigure}
\caption{Median fluency (a) and adequacy percentage (b) per theme after adjusting for stratification by considering sampling weights.}
\label{fig:test}
\vspace*{-12pt}
\end{figure}

We can calculate fluency and adequacy scores for each theme by making use of the sampling weights, accounting for any bias introduced by stratification. Figure~\ref{fig:weighted_aggregated_flu} shows the adjusted median fluency per theme: all have from fair (above 3) to excellent (above 4) fluency, with complex and scientific themes in the lower half. Figure~\ref{fig:weighted_aggregated_ade} shows the adjusted adequacy percentage per theme, ranging from $85.7\%$ to $99.8\%$.

For a bigger-picture view, we calculate the average aggregated fluency and adequacy per partition. This does not consider the sampling weights, as they are not translatable across differently stratified populations. In all aggregated metrics (i.e. mean fluency, median fluency, adequacy percentage, and majority-voted adequacy) \textit{WebLNG\_SEEN} performs the best, followed by \textit{WebNLG\_UNSEEN}, and then \textit{WD\_UNSEEN}. Exact metrics can be seen in Table~\ref{tab:agg_per_partition}. This is in line with how the model was trained and validated. However, the differences are small, signalling excellent generalisation to themes unseen both in training and validation, and also whose provenance is from an entirely different KG.

\begin{table}[ht]
\centering
{
\begin{tabular}{l|l|l|l|}
\cline{2-4}
                                                & WD\_UNSEEN & WebNLG\_UNSEEN & WebNLG\_SEEN \\ \hline
\multicolumn{1}{|l|}{Mean Fluency}              & 3.684      & 3.884          & 3.91         \\ \hline
\multicolumn{1}{|l|}{Median Fluency}            & 3.848      & 4.103          & 4.148        \\ \hline
\multicolumn{1}{|l|}{Adequacy Percentage}       & 80.3\%     & 80.6\%         & 82\%         \\ \hline
\multicolumn{1}{|l|}{Majority-Adequate Perc.}   & 92.5\%     & 92.8\%         & 93.1\%       \\ \hline
\multicolumn{1}{|l|}{Fluency Scores Agreement}  & 0.466761   & 0.508015       & 0.496089     \\ \hline
\multicolumn{1}{|l|}{Adequacy Scores Agreement} & 0.659174   & 0.649527       & 0.654175     \\ \hline
\end{tabular}
}
\vspace*{10pt}
\caption{Aggregated scores and agreement per partition. Mean fluency, median fluency and adequacy percentage were averaged. Majority-Adequate Perc. is the percentage of claims whose majority-voted adequacy score was \textit{Yes}.}
\label{tab:agg_per_partition}
\vspace*{-24pt}
\end{table}

We calculate the agreement for each theme and partition. All themes show agreement above $0.4$ on the fluency task, and above $0.6$ on the adequacy task. Fluency and adequacy agreement metrics per theme have a substantial correlation ($0.63$ Pearson correlation). Agreement did not vary substantially between partitions (see Table~\ref{tab:agg_per_partition}), showing that whether or not the model was trained or validated on a partition did not impact the workers' abilities to judge it.
\section{Verbalisation Model} 
\label{sec:verbalisation}

Our dataset relies on a pre-trained and fine-tuned data verbalisation model for its text alignment. In this section, we describe the model we have chosen and all reasons for it, as well as its training approach and hyperparameters used. We finish by evaluating its fitness for use with examples from our dataset.

\subsection{Approach, Training, and Validation}

Many state of the art KG data verbalisation models take the graph structure into consideration~\cite{trisedya-etal-2018-gtr,zhao-etal-2020-bridging,ijcai2020-419}. GTR-LSTM~\cite{trisedya-etal-2018-gtr} and DualEnc~\cite{zhao-etal-2020-bridging} both encode the graph by combining graph neural networks and recurrent sequential neural networks. Working with single-claim data, we do not need to maintain the graph's structure. Large pre-trained language models have achieved state of the art results when fine-tuned and evaluated on WebNLG~\cite{ribeiro-etal-2021-investigating,harkous-etal-2020-text,guo-etal-2020-2}, mainly the T5~\cite{exploring2019raffel}; they can disregard most structure and can be applied to one or many claims at a time. Hence, we utilise the T5 (base version) as our verbalisation model, following training and evaluation methods from these works.

The T5 converts input text into output text based on a given task, such as summarisation, specified through natural language as a prefix to the input. It can also learn new tasks by being fine-tuned with new data and a new prefix~\cite{exploring2019raffel}. Our model has been fine-tuned on WebNLG\cite{gardent2017creating}. The \textit{SEEN} partition is used for both training and validation/testing, while the \textit{UNSEEN} partition is kept for testing only. We follow the training setup from Ribeiro et al.~\cite{ribeiro-etal-2021-investigating} by specifying a new prefix ``translate from Graph to Text'' and adding three new tokens ($\langle H \rangle$, $\langle R \rangle$, and $\langle T \rangle$) that precede the claim's subject, predicate, and object, respectively.

Each entry in the training data consists of a set aligning multiple triples to multiple sentences. We train the model by concatenating all triples in the set in a random order, marked with the new tokens, and choosing one of the verbalisations at random (as long as they were tagged as \textit{good} by WebNLG). 

Some of the hyperparameters used in the model were: a $3e-5$ learning rate on a linear scheduler with no warmup steps, an Adam~\cite{Kingma2015AdamAM} optimiser with no weight decay and $1e-8$ epsilon, a cross-entropy loss function, 100 epochs of training with early stopping patience of 15, batch size of four for both training and evaluation, and text generation was done via beam search with 3 beams. The T5 was obtained from HuggingFace~\cite{wolf-etal-2020-transformers} and trained with PyTorch Lightning~\cite{pytorchlight}. A full list of hyperparameters is available in this paper's GitHub repository.

We benchmarked our model in two ways. Firstly, by evaluating aggregated crowdsourced metrics, covered in Section~\ref{sec:data_analysis}. Secondly, by calculating the BLEU score~\cite{bleu} for both the \textit{SEEN} and \textit{UNSEEN} partitions, to make sure our model matches current state of the art~\cite{ribeiro-etal-2021-investigating}. BLEU scores were calculated with Moses~\cite{koehn2007moses}. The model achieved $65.51$ on \textit{SEEN}, $51.71$ on \textit{UNSEEN}, and $59.41$ on \textit{ALL}.

\subsection{Evaluation and Output Examples}

Comparisons between the model employed by us (T5-base), other language models, and other mentioned approaches on verbalising the WebNLG dataset have been carried by other authors~\cite{ribeiro-etal-2021-investigating,bosc-etal-2016-dart} and is out of the scope of our paper. To evaluate our model's fitness of use, we will now look at cases in which it performed well and poorly for fluency or adequacy. This allows us to understand how the model could improve, and in which situations its use is and is not effective. Table~\ref{tab:examples} shows a few examples of the model's inputs and outputs.

\begin{table}[ht]
\centering
{
\begin{tabular}{|lll|lll}
\cline{1-3}
\multicolumn{3}{|c|}{Claim}                                                                                                                                                                                                                                     &                                                                                                                                            &                          &                          \\ \hline
\multicolumn{1}{|l|}{Subject}                                                         & \multicolumn{1}{l|}{Predicate}                                                               & Object                                                                   & \multicolumn{1}{l|}{Verbalisation}                                                                                                         & \multicolumn{1}{l|}{Flu.}   & \multicolumn{1}{l|}{Ade.}   \\ \hline
\multicolumn{1}{|l|}{\begin{tabular}[c]{@{}l@{}}Eve Towards\\ the Light\end{tabular}} & \multicolumn{1}{l|}{height}                                                                  & 268 centimetre                                                           & \multicolumn{1}{l|}{\begin{tabular}[c]{@{}l@{}}Eve Towards the Light\\ is 268 centimetres high\end{tabular}}                               & \multicolumn{1}{l|}{2.0} & \multicolumn{1}{l|}{1.0}   \\ \hline
\multicolumn{1}{|l|}{Bydgoszcz}                                                       & \multicolumn{1}{l|}{flag}                                                                    & flag of Bydgoszcz                                                        & \multicolumn{1}{l|}{\begin{tabular}[c]{@{}l@{}}The flag of Bydgoszcz\\ is the flag of Bydgoszcz\end{tabular}}                              & \multicolumn{1}{l|}{0.0} & \multicolumn{1}{l|}{1.0}   \\ \hline
\multicolumn{1}{|l|}{Marawi}                                                          & \multicolumn{1}{l|}{\begin{tabular}[c]{@{}l@{}}Köppen climate\\ classification\end{tabular}} & Tropical climate                                                         & \multicolumn{1}{l|}{\begin{tabular}[c]{@{}l@{}}city of the Philippi-\\ nes and capital of the\\ province of Lanao del\\ Sur\end{tabular}}  & \multicolumn{1}{l|}{4.0} & \multicolumn{1}{l|}{0.4} \\ \hline
\multicolumn{1}{|l|}{Decius}                                                          & \multicolumn{1}{l|}{child}                                                                   & Hostilian                                                                & \multicolumn{1}{l|}{\begin{tabular}[c]{@{}l@{}}Decius is a\\ child of Hostilian\end{tabular}}                                              & \multicolumn{1}{l|}{4.0} & \multicolumn{1}{l|}{0.4} \\ \hline
\multicolumn{1}{|l|}{\begin{tabular}[c]{@{}l@{}}UCAC2\\ 28288086\end{tabular}}        & \multicolumn{1}{l|}{surface gravity}                                                         & \begin{tabular}[c]{@{}l@{}}9 centimetre per\\ square second\end{tabular} & \multicolumn{1}{l|}{\begin{tabular}[c]{@{}l@{}}UCAC2 28288086\\ has a surface gravity\\ of 9 centimetres\\ per square second\end{tabular}} & \multicolumn{1}{l|}{4.0} & \multicolumn{1}{l|}{1.0} \\ \hline
\multicolumn{1}{|l|}{Casa Batlló}                                                     & \multicolumn{1}{l|}{fabrication method}                                                      & Trencadís                                                                & \multicolumn{1}{l|}{\begin{tabular}[c]{@{}l@{}}Casa Batlló is made\\ using the Trencadís\\ method\end{tabular}}                            & \multicolumn{1}{l|}{5.0} & \multicolumn{1}{l|}{0.8} \\ \hline
\end{tabular}
}
\vspace*{10pt}
\caption{Examples of claims verbalised by the model. On the left, we see the claims and their components' labels. On the right, the verbalisations and their aggregated metrics of fluency (median) and adequacy (adequacy percentage).}
\vspace*{-24pt}
\label{tab:examples}
\end{table}

We consider a low fluency score to be under 3, when grammatical errors are not minor or text is not comprehensible. Out of over 1.4k annotated claim\hyp{}verbalisation pairs, 55 had low fluency. A considerable amount of them ($41\%$) suffer due to subject or object labels having complex syntaxes, such as IUPAC chemical nomenclatures, names of astronomical bodies, and full titles of scientific papers. These are challenging both for the model and for workers with no context or knowledge of how to use these names in a sentence. This potential misinterpretation is evidenced by $38\%$ of all low-fluency verbalisations being simply misinterpreted by the crowd; the sentences are fluent, but have non-trivial or non-English terms that throw workers off e.g. ``Eve Towards the Light is 268 centimetres high'', which describes a painting.
Around a third ($32\%$) of cases were the model's fault, either by failure to structure the predicate or by corrupting or inverting subject and object labels. However, $21\%$ of cases could be solved by improving predicates and entity labels, or rethinking how information is stored in Wikidata; some predicates are vague or depend on qualifiers to make complete sense e.g. \texttt{inception} and \texttt{different from}, and some claims have redundant data e.g. ``The flag of Bydgoszcz is the flag of Bydgoszcz''. 

Low adequacy is when the majority-voted option for adequacy was \textit{No}. This corresponds to 78 verbalisations. Almost half ($46.15\%$) consists of claims either for which the model could not properly structure the predicate e.g. ``Köppen climate classification'' or for which subject and predicate had complex or non-English labels. Over a third ($38.4\%$) of these were adequate claims that were misunderstood by the crowd e.g. ``Craig-y-llyn is designated as a Site of Special Scientific Interest''. Somewhat often ($17.9\%$), vague predicates and badly written labels were also to blame. Lastly, the model would sometimes ($11.5\%$) either shift subject with object, infer information not seen in the claim (delusions), or translate words between English and German (one of T5's other learned tasks).

These cases show us that the verbalisation model can be improved either by design or through data curation. For instance, predicates that rely on qualifiers can have that information communicated to the model if the model can properly link them to specific components of the claim. We can avoid inversion of subject and object by adding direction either on the predicate labels (e.g. \textit{child} to \textit{has child}) or through the model's encoding. We managed to help the model understand certain predicates and entities by using alternative labels (e.g. \textit{conflict} to \textit{participated in conflict}), but which aliases to use is very context dependant. 

Some issues are less trivial to address. Entities with syntactically complex labels hardly have simpler aliases. Vague predicates might be solved by using aliases, but this is extremely context-sensitive, and there might be good reasons why these predicates unite multiple senses under a common abstraction (e.g. \texttt{facet of} and \texttt{inception}. Finally, redundant information can emerge from Wikidata's predicates. For instance, an entity exists for the city of Bydgoszcz, and another for its flag, containing information such as its appearance. They are linked by the \texttt{flag} predicate. This makes ontological sense, but no verbal sense, as one would express this relationship as either ``Bydgoszcz has a flag'' or ``Bydgoszcz's flag is Bydgoszcz's flag''; this is either redundant or inadequate.
\section{Addressing Review Criteria} 
\label{sec:addressing_review_criteria}

Here, we further strengthen the argument that the resources presented are not only of interest to Semantic Web researchers, but have a provable claim to adoption by them and the Wikidata research community. These resources support a line of research by the same authors on the quality of Wikidata references, which proposes crowdsourcing and computational methods to assess different dimensions of reference quality. The first part of the work assessed reference accessibility, relevance and authoritativeness based on features that are not directly related to the content of the reference themselves. It has been recently awarded the Wikimedia Research Paper of the Year $2022$, from among $230$ peer-reviewed papers. The judges highlighted the importance of the research problem (reference quality) and the value of the solution proposed, especially in a multilingual setting. WDV directly builds on top of this, by feeding into computational methods that allow us to assess reference quality also in terms of the actual content in the reference source. It has already made possible the authors' efforts towards automated fact verification in Wikidata.

Wikidata recognises references as essential in its own guidelines, stating that ``Wikidata is not a database that stores facts about the world, but a secondary knowledge base that collects and links to references to such knowledge''.\footnote{https://www.wikidata.org/wiki/Help:Statements}. They  promote reference quality assurance efforts, as many open phabricator tickets show.~\footnote{\url{https://phabricator.wikimedia.org/T90881}} \footnote{\url{https://phabricator.wikimedia.org/T156389}} The Wikidata editing community also discusses at length the need for automated techniques for reference quality assessment.\footnote{\url{https://www.wikidata.org/wiki/Property_talk:P1456}} \footnote{\url{https://www.wikidata.org/wiki/Wikidata:Project_chat/Archive/2017/10#Proposal_on_citation_overkill}}

%
\section{Conclusion} 
\label{sec:discussion_and_conclusion}

In this paper, we have presented WDV: a large dataset for the verbalisation of single triple-based claims from Wikidata (a collaborative KG). It directly aligns claims to natural language sentences that aim at being grammatically well-written and transmitting the same meaning. WDV was created to provide a data-to-text resource that covers a wide range of entities, topics, and predicates in Wikidata. More importantly, it does so in a balanced manner, so that specific themes are not overly represented. We also presented and carried an evaluation workflow of the fluency and adequacy of its natural language sentences, concluding that they have very high levels of both metrics.


We believe this dataset constitutes a valuable step towards understanding how to efficiently carry the verbalisation of triple claims from Wikidata and KGs in general. Bridging the gap between labelled triple components and natural language is crucial to implementing downstream NLP tasks in the KG. One such task that can be helped immensely by this resource is the automated fact-checking of KG claims based on the textual information found in the references they cite. Finally, WDV, alongside the annotation workflow we have defined, can promote the evaluation, through a human perspective, of NLG models performances without relying on algorithmic metrics.

\subsubsection*{Acknowledgements} This research received funding from the European Union’s Horizon 2020 research and innovation programme under the Marie Skłodowska-Curie grant agreement no. 812997.
%
%
%
\bibliographystyle{splncs04}
\bibliography{mybibliography}
%
%
%
%
%
\end{document}